\documentclass[journal]{IEEEtran}

\ifCLASSINFOpdf
\else
\fi

\usepackage{amsmath,amssymb,amsfonts}
\usepackage{graphicx}
\usepackage{textcomp}
\usepackage{color}
\usepackage{makecell,multirow,diagbox}
\usepackage{algorithm,algpseudocode}
\usepackage{fancyhdr}
\usepackage{color}
\usepackage{microtype}
\usepackage{url}
\usepackage{multirow}
\usepackage{array}
\usepackage{caption}
\usepackage{subcaption}
\usepackage{verbatim}
\usepackage{tabularx}

\newtheorem{Def}{Definition}
\newtheorem{Theo}{Theorem}

\newtheorem{lem}{Lemma}

\newcommand{\RNum}[1]{\uppercase\expandafter{\romannumeral #1\relax}}
\newcommand{\tabincell}[2]{\begin{tabular}{@{}#1@{}}#2\end{tabular}}

\begin{document}
%
\title{ {Correlated Differential Privacy: Feature Selection in Machine Learning}}
%
%
%

\author{Tao Zhang, Tianqing Zhu$ ^{*} $, Ping Xiong, Huan Huo, Zahir Tari, Wanlei Zhou
\thanks{ 
	Tao Zhang, Huan Huo, Wanlei Zhou are with the School of Computer Science, University of Technology, Sydney,
	Australia. Email: $ \{ $Tao.Zhang-3@student.uts.edu.au, Huan.Huo@uts.edu.au, Wanlei.Zhou@uts.edu.au$ \} $
	
	Tianqing Zhu is with two affiliations: China University of Geosciences, School of Computer Science, Wuhan China 
	and University of Technology Sydney, School of Computer Science. Email: $ \{ $Tianqing.Zhu@uts.edu.au$ \} $
	
	P. Xiong is with the School of Information and Safety Engineering, Zhongnan University of Economics and Law, Wuhan, China.
	Email: $ \{ $pingxiong@znufe.edu.cn$ \} $
	
	Zahir Tari is with the School of Computer Science, CE$ \& $SE discipline, RMIT University, Melbourne, Australia.
	Email: $ \{ $zahir.tari@rmit.edu.au$ \} $
	
}}


%
%

\markboth{Journal of \LaTeX\ Class Files}%
{Shell \MakeLowercase{\textit{et al.}}: Bare Demo of IEEEtran.cls for Journals}
%


\maketitle

\begin{abstract}
	Privacy preserving in machine learning is a crucial issue in industry informatics since data used for training in industries usually contain sensitive information. Existing differentially private machine learning algorithms have not considered the impact of data correlation, which may lead to more privacy leakage than expected in industrial applications.  For example, data collected for traffic monitoring may contain some correlated records due to temporal correlation or user correlation.
	To fill this gap, we propose a correlation reduction scheme with differentially private feature selection considering the issue of privacy loss when data have correlation in machine learning tasks. 
	The proposed scheme involves five steps with the goal of managing the extent of data correlation, preserving the privacy, and supporting accuracy in the prediction results.
	In this way, the impact of data correlation is relieved with the proposed feature selection scheme, and moreover the privacy issue of data correlation in learning is guaranteed. The proposed method can be widely used in machine learning algorithms which provide services in industrial areas.
	Experiments show that the proposed scheme can produce better prediction results with machine learning tasks and fewer mean square errors for data queries compared to existing schemes.

\end{abstract}


%
\IEEEpeerreviewmaketitle

\section{Introduction}

\IEEEPARstart{C}{urrently}, machine learning becomes an indispensable tool to provide services for human beings in industrial applications, such as Internet of Things (IoT) \cite{bibitem Shanthamallu2017} and smart cities \cite{bibitem Hashem2016}. 
One main data source used for machine learning in industry is from human’s activities. For example, human’s data are often collected via smart phones and these data are analyzed to provide some services in smart cities, such as traffic monitoring \cite{bibitem Yin2017} and smart health \cite{bibitem Solanas2014}.  Data collected from human usually contain some sensitive information, as the location information and health data in above examples. When these data are used for machine learning, individual privacy can be leaked  \cite{bibitem Fung}.


As a popular technique for privacy preserving, differential privacy was first proposed by Dwork et al. \cite{bibitem Dwork2006}. Since then, differential privacy has attracted considerable attention because it provides a rigorous mathematical framework for preserving privacy. 
Recently, differential privacy is widely used to protect the privacy in industrial informatics, such as location privacy protection \cite{bibitem Yin2017}, \cite{bibitem Yang2018}, smart grids \cite{bibitem Lyu2018}, \cite{bibitem Liu2019} and multi-agent systems \cite{bibitem Ye2019}. 

Much work has addressed the privacy issue in machine learning with differential privacy.  Chaudhuri provided an output perturbation where the model was trained and then the noise was added to the output \cite{bibitem Chaudhuri2009} and objective perturbation mechanism  where a carefully designed linear perturbation item was added to the original loss function \cite{bibitem Chaudhuri2011}. \cite{bibitem Song2013} derived differentially private stochastic gradient descent mechanisms and tested them empirically in logistic regression. \cite{bibitem Abadi2016} proposed a differentially private deep learning algorithms which was based on a differentially private version of stochastic gradient descent. \cite{bibitem Zhu2018-1} studied the differentially private publishing model. However, previous works have not considered the data correlation when designing differentially private machine learning algorithms.

In the definition of differential privacy, data in a dataset are assumed to be independent. This is a somewhat faulty assumption since data in industrial applications are always correlated beginning from when the data is first generated, such as temporal datasets in monitoring systems. Intuitively, when some of the records in a dataset are correlated, deleting one record may have a great impact on the other records, which could reveal more information to an adversary than expected. Kifer and Machanavajjhala’s study on data correlation \cite{bibitem Kifer2011} confirms this observation, and the finding has launched a new stream of research on how to preserve privacy in correlated datasets. \cite{bibitem Zhu2015}, \cite{bibitem Zhu2018}  introduced correlation parameters to describe data correlation.  Correlation models were proposed to model data correlation, such as the Gaussian correlation model in \cite{bibitem Yang2015}, \cite{bibitem Chen2017} and Markov chain models in \cite{bibitem Cao2017}. Also, \cite{bibitem Kifer2014} designed a second privacy framework, called Pufferfish, which is flexible and can provide a privacy guarantee for various data sharing needs.

Correlated data used for industrial applications can also disclose more privacy information in machine learning algorithms when applying differential privacy.  Previous methods do not always guarantee good performance because data correlation is not always easy to capture or describe accurately in the real world.
Unlike previous studies, the proposed scheme correlation reduction based on feature selection (CR-FS)  reduces data correlation and can be applied to both data analysis and data publishing, which provides a widely used applications in industries. Feature selection is a key method in machine learning for choosing the features that are crucial to predicting a result \cite{bibitem Chandrashekar}. It is used to reduce overfitting, but can also be used to reduce data correlation across an entire dataset. 


Overall, the contributions of this paper can be summarized as follows:
\begin{itemize}
	\item 1) We proposed a differentially private feature selection based on feature importance. The proposed method can select features privately, while retaining a desirable data utility.
	
	\item 2) We propose a correlation reduction scheme based on feature selection to reduce data correlation in correlated datasets. This helps to reduce the correlated sensitivity when implementing differentially private machine learning algorithms, and thus improves data utility.
	
	\item 3) Experiments validate the effectiveness of our proposed feature selection scheme. The results show improved data utility for both data analysis and data publishing.
\end{itemize}


\section{Preliminaries}

\subsection{Differential privacy}
Differential privacy is a rigorous privacy model \cite{bibitem Dwork2014}. In brief, given two datasets $ D $ and $D^{'}$ that contains a set of records, these are referred as neighboring datasets when they differ in one record.
A query $ Q $ is a function that maps the record $ r \in D $  into  outputs $ Q(D) \in \mathcal{R} $,
where $\mathcal{R}$ is the whole set of outputs. 

\begin{Def}
	($\epsilon$-Differential privacy) A randomized algorithm $M $ satisfies
	\emph{$ \epsilon $-differential privacy} if for any pair of datasets, say $ D $  and $ D^{'} $, and for any possible outcome  $ Q(D) \in R $, we have
	\begin{equation} Pr[\mathcal{M}(D) \in  R] \leq exp(\epsilon) \cdot Pr[\mathcal{M} (D^{'}) \in  R  ]  \label{eq}  \tag{1}\end{equation}
	
	\noindent where $ \epsilon $ refers to the privacy budget that controls the privacy level of the mechanism
	$ \mathcal{M} $. The lower $\epsilon$ represents the higher privacy level.
\end{Def}

\begin{Def}
	(Sensitivity) For a query $ Q:D\xrightarrow{} \mathcal{R} $, and neighboring datasets, the sensitivity of $ Q $ is defined as
	\begin{equation} \Delta f= \max \limits_{D,D^{'} } ||Q(D)-Q(D^{'})||_1 \label{eq} \tag{2}\end{equation} 
	Sensitivity describes the maximal difference between neighboring datasets, which is only related to the type of query $ Q $.
\end{Def}


\begin{Def}
	(Laplace mechanism) For any query $ Q $: $ D\xrightarrow{}\mathcal{R} $ over the database $ D $,
	the following mechanism provides $\epsilon$-differential privacy if
	\begin{equation} \mathcal{M}(D)=Q(D)+Laplace(\Delta / \epsilon) \label{eq} \tag{3}\end{equation}
	
\end{Def}
The Laplace noise is denoted as $ Laplace(\cdot) $ and is drawn from a Laplace distribution with the
probability density function $ p(x| \lambda)=\frac{1}{2 \lambda}e^{-|x|/ \lambda} $, where $ \lambda $ relate to the sensitivity and the privacy budget.

\begin{Theo}
	Sequential composition:  Suppose that a set of privacy mechanisms  $\mathcal{M}$=\{${\mathcal{M}_{1},...,\mathcal{M}_{m}}$\}, gives $ \epsilon_i $ differential privacy ($ i=1,2...,m $), and these mechanisms are sequentially performed on a dataset. $ \mathcal{M} $ will provides  $ (\sum_i\epsilon_i )$-differential privacy for this dataset.
\end{Theo}

\subsection{Feature selection}
Feature selection is a method for selecting the attributes in a dataset (such as columns in tabular data) that are most relevant to the prediction \cite{Liu2005}. In other words, feature selection largely acts as a filter that sifts out features that are less useful to solving a problem. With feature selection, both the efficiency and the accuracy of the predicted results can be improved.

In this paper, we adopt feature importance to select features. Feature importance is a method of ranking features based on random forests.  Feature importance is measured according to the mean decrease in impurity, which is defined as the total decrease in node impurity averaged over the forest. This score can be computed automatically for each feature after training and scaling the results so that the sum of importance for all features is equal to 1. One strength of the random forest is that it is easy to measure which features are relatively more important to the results. With this method, we are able to select the most important features in the dataset.
\section{Example of the traffic monitoring }
In this section, we present the issue of correlated data in differential privacy with a detailed industrial example of traffic monitoring and show how correlated data can degrade the level of privacy in industry applications.

The traffic monitoring is one of most used technologies in smart cities. 
User's location information in a region are collected by a trusted server and the aggregate information of the dataset (i.e., the counts of users at each location) is continuously released to the public. Some users in the region may have a form of social relationship – perhaps family members. In this case, some users may have the same location information during some time and hence the records of users' information can be correlated in the dataset.


As shown in Table \RNum{1}, the user's locations are recorded at different time points. It is assumed that users only appear in one location at each time point, and it is observed that $ user_{1} $  and $ user_{2} $  take the same route from time point $ t=1 $ to $ t=4 $ (they may have social relationships).  In this case, if one were to change the location of $ user_{1} $, the location of $ user_{2} $ would also change. In this way, the records for $ user_{1} $ and the records for $ user_{2} $ are correlated.

\begin{table}[!htbp]
	\small
	\centering
	\caption {  Users' locations at different times}
	\begin{tabular}{|c|c|c|c|c|}
		\hline
		\diagbox{user}{t}&{1}&{2}&{3}&{4}\\ 
		\hline 
		$ u_{1} $&$ loc_{2} $&$ loc_{2} $&$ loc_{3} $&$ loc_{4} $\\
		\hline
		$ u_{2} $&$ loc_{2} $&$ loc_{2} $&$ loc_{3} $&$ loc_{4} $\\
		\hline
		$ u_{3} $&$ loc_{1} $&$ loc_{4} $&$ loc_{5} $&$ loc_{2} $\\
		\hline
		$ u_{4} $&$ loc_{4} $&$ loc_{5} $&$ loc_{2} $&$ loc_{5} $\\
		\hline
	\end{tabular}
\end{table}

\begin{table}[!htbp]
	\small
	\centering
	\caption { The sum counts of users' locations}
	\begin{tabular}{|c|c|c|c|c|}
		\hline
		\diagbox{loc}{t}&{1}&{2}&{3}&{4}\\ 
		\hline 
		$ loc_{1} $&1&0&0&0\\
		\hline
		$ loc_{2} $&2&2&1&1\\
		\hline
		$ loc_{3} $&0&0&2&0\\
		\hline
		$ loc_{4} $&1&1&0&2\\
		\hline
		$ loc_{5} $&0&1&1&1\\
		\hline
	\end{tabular}
\end{table}

Table \RNum{2} shows that the some counts  at different locations are always 2. In terms of the Laplace mechanism, adding the amount of $ Lap(1/\epsilon) $ noise to perturb each count in Table 2 can achieve $ \epsilon $-DP at each time point. However, the expected privacy guarantee may breach with correlated records in the dataset. With background information of who has the relationship in a certain region, an attack can infer the location information of $ user_1 $ and $ user_2 $ at different time points. 
 Consequently, after releasing private count of user's locations, the location information of $ user_1 $ and $ user_2 $ may not be $ \epsilon $- differentially private as expected. Instead, it is $ 2\epsilon $-differentially private since changing one user's location will change the count 2.

In summary, this example shows that correlated data in a dataset will disclose more information than expected when these data are used for machine learning algorithms in industrial applications. Essentially, adding more noise to a correlated dataset is a way to guarantee differential privacy. Such a case reveals the level of challenge in industries when dealing with correlated data in situations where differential privacy must be satisfied, but high-quality query results must be maintained.

\section{The extent of data correlation}
\subsection{Correlated degree }


Inspired by  \cite{bibitem Zhu2015}, we have incorporated the notion of correlated degree $ \theta_{ij} \in [-1,1] $ to denote the extent of correlation between record $i $ and record $ j $. When $ |\theta_{ij}| >0 $, record $i$ and record $j$ have a positive correlation and vice versa. When $ |\theta_{ij}| =1 $, record $ i $ and record $ j $ are fully correlated and When  $ \theta_{ij} =0 $, there is no relationship. When there are a number of $ l $ records in a dataset, it is possible to list the relationship for all records and form a correlated degree matrix $ \Lambda $.
\begin{equation}
\Lambda=
\left(
\begin{matrix}
\theta_{11}     & \theta_{12}      & \cdots & \theta_{1l}      \\
\theta_{21}     & \theta_{22}      & \cdots & \theta_{21}      \\
\vdots & \vdots & \ddots & \vdots \\
\theta_{l1}      & \theta_{l2}      & \cdots & \theta_{ll}      \\
\end{matrix}
\right) \tag{5}
\end{equation}
A threshold $ \theta_0 $ is defined so as to select strongly correlated records. For a given $ \theta_0 $, the value of the correlated degree is
\begin{equation}
\theta_{ij}=
\begin{cases}\theta_{ij},\quad \ \ &   \theta_{ij} \geq \theta_0,\\   0, \quad \ \ &  \theta_{ij} < \theta_0, \end{cases} \tag{6}\end{equation}
A correlated degree matrix can describe the correlations of the whole dataset and, once analyzed, the curator will hold all knowledge of the data correlations. 
Data privacy can still be protected, even when the adversary is privy to the entire correlated degree matrix, if enough noise is added to mask the highest impact of deleting one record using correlated differential privacy.

\begin{figure*}[htbp]
	\centering
	\includegraphics [scale=0.7] {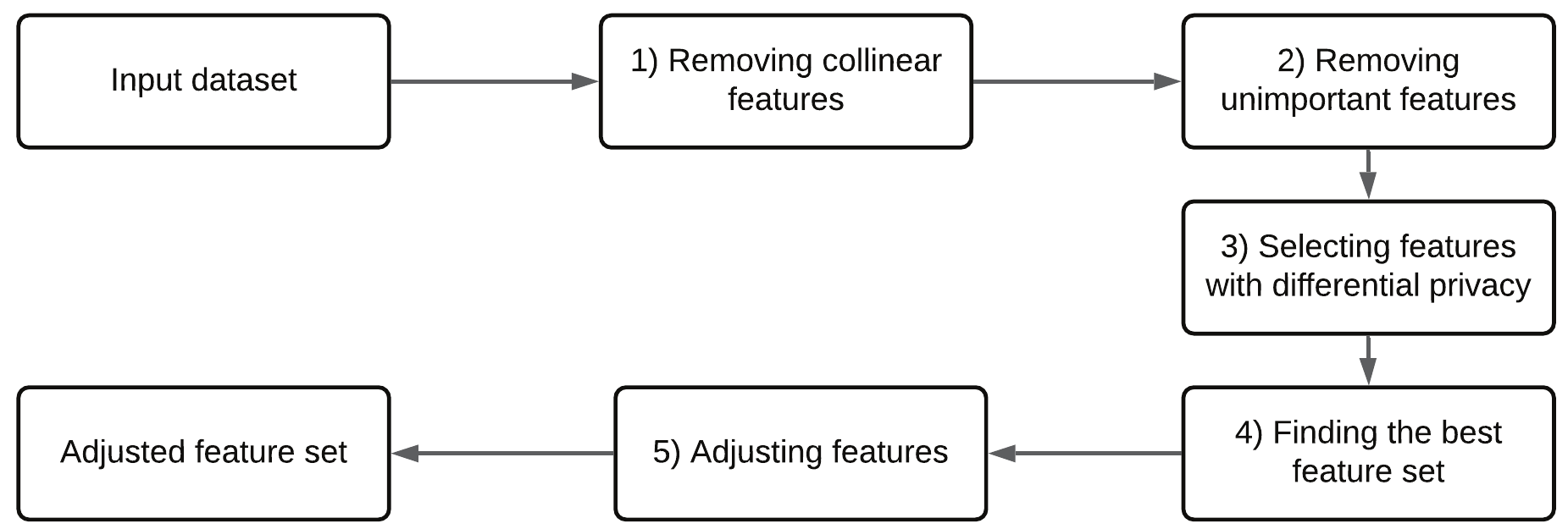}
	\caption{The process of feature selection}
\end{figure*}

\subsection{Correlated sensitivity }
Global sensitivity can only measure the maximal number of correlated records but does not consider the extent of the data correlation. Hence, the notion of correlated sensitivity is introduced to measure the extent of the impact on other records from changing one record. As mentioned earlier, global sensitivity adds extra noise by simply multiplying the maximal number of correlated records. Whereas, correlated sensitivity is able to model the correlations in a more exact way.

\begin{Def}
	(Correlated sensitivity) For a query $ Q $, correlated sensitivity is based on the correlated degree and the number of correlated records, which is defined as
	\begin{equation} \Delta CS_{q}=\max \limits_{i \in q} \sum_{j=0}^{l} |\theta_{ij}| \{	 \lVert (Q(D^{j})-Q(D^{-j}) \rVert_{1}    \} \tag{7} \end{equation}

\end{Def}
where $ q $ is the set of records in a dataset,  and $ \theta_{ij} $ is the correlated degree between record $ i $ and record $ j $. $ D_{j} $ and $ D_{-j} $ are neighboring datasets that differ by record $ j $. Correlated sensitivity lists all the sensitivity of records with the query $ Q $. With correlated sensitivity, the maximal effect on all records of a dataset can be measured when one record is deleted. 
For any query $ Q $, the perturbed answer is calibrated with the equation,
\begin{equation}
\hat{Q}(D)=Q(D)+Laplace(\frac{\Delta CS_{q}}{\epsilon}) \tag{8}
\end{equation}
For any query $ Q $, the correlated sensitivity is smaller than the global sensitivity. The global sensitivity is denoted as $ \Delta GS_{q}=\max \limits_{i \in q} \sum_{j=0}^{k}  \{	 k\lVert (Q(D^{j})-Q(D^{-j}) \rVert_{1}    \} $, where $ k $ denotes the number of correlated records.
Since we use the correlated degree $ \theta_{ij} \in [-1,1] $ to describe the extent of data correlation, the correlated sensitivity is no larger than the global sensitivity.

We note that the correlated degree $ \theta_{ij} $ is related to every feature in record $ i $ and record $ j $. When deleting features in the dataset, the extent of correlation between record $ i $ and record $ j $ will also be changed. Thus, after describing the extent of data correlation in a dataset, we use feature selection to reduce data correlation.

\section{Correlation reduction based on feature selection}
\subsection {Overview of the method }
In our method, we select features in terms of three principles: 1) the accuracy of training results; 2) the privacy of feature selection; 3) the reduction of the data correlation. As Fig. 1 shows,  the proposed scheme CR-FS involves five steps: 1) removing collinear features; 2) removing unimportant features; 3) choosing features with differential privacy;
 4) obtaining the Best feature set $  \mathcal{B} $; and 5) adjusting the features that can reduce data correlation within the dataset. Each of these methods is described in detail in the following sections.

\subsection{The proposed CR-FS scheme} 
Following traditional feature selection, we propose the algorithm \RNum{1} that selects features with differential privacy. 	
For a given dataset, feature selection is a crucial step before executing a machine learning algorithm, especially with high-dimensional datasets. 
 Additionally, retaining more features typically leads to a higher degree of data correlation, which, with differential privacy, negatively impacts the privacy level. Hence, our goal is to select a subset of features with relatively lower levels of data correlation while maintaining good utility for data publishing and analysis.

\begin{algorithm}[t]
	\small 
	\caption{ Differentially private feature selection scheme} 
	\hspace*{0.02in} {\bf Input}: 
	Dataset, $ T_{cf}, T_{fi}, T_{mv}  $, $ \epsilon_{1} $;\\
	\hspace*{0.02in} {\bf Output:} 
	Best feature set $ \mathcal{B} $, Adjusted feature set $ \mathcal{A} $;
	\begin{algorithmic}[1] 
		\State Calculate  feature collinearity $ \rho_{f_{m},f_{n}}=\frac{E[(f_{m}-\mu_{f_{m}})(f_{n}-\mu_{f_{n}})]}{\sigma_{f_{m}}\sigma_{f_{n}}} $; /* Step 1 */
		\If {$\rho_{f_{m},f_{n}}    \leq T_{cf} $   }   
		\State Remove $ f_m $ or $ f_n $;
		\EndIf
		
		\State Remove unimportant features with $ T_{fi} $; /* Step 2 */
		\State Remove missing values with $T_{mv}$
		
		\State Calculate the $ fim_n $ of features by Random forest; /* Step 3 */
		\State Calculate the sensitivity $ \Delta fim $ according to Equation (11);
		\For {$fim_{n}$; n=1,2,...,N:}
		\State Add Laplace noise $ \hat{fim_{n}}=fim_{n}+Lap(\frac{\Delta fim_{q}}{\epsilon_{1}})  $;
		\EndFor
		\State Do the normalization $ {fim_n}= \hat{fim_{n}}/ \sum_{n=1}^{N}\hat{fim_{n}} $;

		\For {i=1,2,...,n:}  /* Step 4 */
		\State Delete features one by one according to the sequence of feature importance and
		calculate the prediction;
		\EndFor
		\State Find the Best feature set : $\mathcal{B}=\{f_1, f_2,...f_k\}$ and Adjusted feature set: $\mathcal{A}=\{f_{k+1},...,f_n\}$; 
		\State Add or delete features from Adjust feature set $ \mathcal{A} $ according to
		$ \mathbf{algorithm \;2} $;
	\end{algorithmic}
\end{algorithm}

\subsubsection{Removing collinear features}
The first step is to filter out the collinear features that can decrease generalization performance on the test set due to less model interpretability and high variance. Usually, the extent of collinearity between features is calculated by the absolute magnitude of the Pearson’s correlation coefficient. The calculation of Pearson's correlation coefficient is 
\begin{equation}
\rho_{f_{m},f_{n}}=\frac{E[(f_{m}-\mu_{f_{m}})(f_{n}-\mu_{f_{n}})]}{\sigma_{f_{m}}\sigma_{f_{n}}} \tag{9}
\end{equation}
Where $ f_{m} $ and $ f_{n} $ are two random features in the dataset; $ \mu_{f_{m}} $ and $ \mu_{f_{n}} $ are the mean of $ f_{m} $ and $ f_{n} $; $ \sigma_{f_{m}} $ and $ \sigma_{f_{m}} $ are the standard deviation of feature $ f_{m} $ and $ f_{n} $.
In our scheme, we set a threshold of $ T_{cf} \in [0,1] $ to identify collinear features and remove the features with a collinearity of greater than $ T_{cf} $.

\subsubsection{Removing unimportant features}
The second step is to remove unimportant features, including 1) features of zero importance and features of low importance; 2) features with a high percentage of missing values; and 3) features with a single value.  Zero and low importance features can be identified using the feature importance threshold, denoted as $ T_{fi} \in [0,1] $. Features with an importance value of lower than  $ T_{fi} $ will be removed. 
The threshold for missing values is defined as $ T_{mv} \in [0,1] $, and features with a percentage of missing values greater than $ T_{mv}  $  will be removed. 

\subsubsection{Choosing features with differential privacy}
We adopt feature importance $ fim $ in Random forest to calculate the feature weight for each feature.  Neighboring data is obtained when record $ r_i $ is deleted, the feature importance can be calculated by Random forest and the feature importance $ fim^{i}_{1}, fim^{i}_{2},...,fim^{i}_{N} $ are sorted in an increasing order. Based on this, we introduced the notion of record sensitivity of feature importance.

\begin{Def} 
	(Record sensitivity of feature importance) For a query $ Q $, the record sensitivity of feature importance of $ r_i $  can be defined as,
\end{Def}

 \begin{equation}
  \Delta fim_{i}= ||fim^{i}_{N} -fim^{i}_{1}||_1  \tag{10}
 \end{equation}

\begin{Def} (Sensitivity of feature importance)
For a query $ Q $, the sensitivity of feature importance is determined by the maximal record sensitivity of feature importance,	
\end{Def} 
\begin{equation}
\Delta fim_{q} = \max \limits_{i \in q} (\Delta fim_{i} )\leq 1 \tag{11}
\end{equation}
where $ q $ is a set of records related to a query $ Q $. 
It is easy to know the sensitivity of feature importance is $ \Delta fim_{q} \leq 1 $, since the range of feature importance is from 0 to 1. 
We apply Laplace mechanism to add noise to the feature importance. The perturbed feature importance can be denoted as,
\begin{equation}
\hat{fim_{n}}=fim_{n}+Lap(\frac{\Delta fim_{q} }{\epsilon}) \tag{12}
\end{equation}

Since the sum of the feature importance $ \sum_{n=1}^{N} fim_{n}=1 $, we normalize the perturbed feature importance as follow,
\begin{equation}
{fim_n}= \hat{fim_{n}}/ \sum_{n=1}^{N}\hat{fim_{n}} \tag{13}
\end{equation}

The new sequence of feature importance can be denoted as $ {fim_{1}} < {fim_{2}}<...<{fim_{n}} $.

\subsubsection{Finding the best feature set}
The third step is to find the best feature set. The Best feature set $ \mathcal{B} $ contains the features that will produce the best prediction results by the machine learning algorithm. In our method, the less important features are deleted one by one in the order of feature importance until the best chance of accurate predictions is achieved.
Practically, finding Best feature set with this method demands far less computational overhead than other methods.
The features that have not been selected for Best feature set are stored as the Adjusted feature set. These features can be used later for a tradeoff between utility and privacy. The Best feature set $ \mathcal{B} $ can be denoted as $\mathcal{B} =\{f_{1}, f_{2},...,f_{k} \} $ and the Adjusted feature set $ \mathcal{A} $ can be denoted as $ \{f_{k+1}, f_{k+2},...,f_{N}\} $.

\subsubsection{Adjusting feature scheme}
The final step is to adjust some features based on the Best feature set $\mathcal{B}   $ in order to reduce data correlation over the whole dataset, as a way to balance the tradeoff between utility and correlated sensitivity.
Basically, the correlated sensitivity of a dataset is irrelevant to the number of features. This means that more features of a dataset may have a lower correlated sensitivity and less features may have a higher correlated sensitivity.  Best feature set $ \mathcal{B} $  can achieve a good data utility without privacy guarantee, yet it may have a higher correlated sensitivity  and  a high correlated sensitivity  has a huge impact on utility for data publishing and data  analysis. In other words, if the goal is to generate a differentially private dataset with good data utility, the process of feature selection should also consider correlated sensitivity. 

\begin{algorithm}[!h]
	\small 
	\caption{ Adjusted feature selection scheme } 
	\hspace*{0.02in} {\bf Input:} 
	Best feature set $ \mathcal{B} $, Adjusted feature set $ \mathcal{A} $, $ \epsilon_2, \theta_0 $;\\
	\hspace*{0.02in} {\bf Output:} 
	Adjusted feature set  $ \mathcal{A} $;
	\begin{algorithmic}[1]
		\For {$ f_i \subseteq \{f_{k+1},...,f_N\} $:}
		\State Add features to the Best feature set $ \mathcal{B} $ from the Adjusted feature set $ \mathcal{A} $;
		\State Calculate the correlated sensitivity of new datasets $\Delta CS_{q}=\max \limits_{i \in q} \sum_{j=0}^{l} |\theta_{ij}| \{	 \lVert (Q(D^{j})-Q(D^{-j}) \rVert_{1} $;
		\State Add Laplace noise $ Lap=\frac{\Delta CS_{q}}{\epsilon_2} $;
		\State Train the dataset and get the predicted result;
		\EndFor
		\State Obtain the Adjusted feature set $ \mathcal{A}_1 $ that has the best performance;
		\For {$ f_i \subseteq \{f_1,...,f_k\} $:}
		\State Delete features from the Best feature set $ \mathcal{B} $ one by one;
		\State Calculate the correlated sensitivity of new datasets $\Delta CS_{q}=\max \limits_{i \in q} \sum_{j=0}^{l} |\theta_{ij}| \{	 \lVert (Q(D^{j})-Q(D^{-j}) \rVert_{1} $;
		\State Add Laplace noise $ Lap=\frac{\Delta CS_{q}}{\epsilon_2} $;
		\State Train the dataset and get the predicted result;
		\EndFor
		\State Obtain the Adjusted feature set $ \mathcal{A}_2 $ that has the best prediction;
		\If { $s ( \mathcal{A}_1 ) \geq  s( \mathcal{A}_2 ) $ }
		\State  $ \mathcal{A}_1 $ is the Adjusted feature set $ \mathcal{A}  $;
		\Else
		
		\State $ \mathcal{A}_2 $ is the Adjusted feature set $ \mathcal{A}  $;
		\EndIf		
	\end{algorithmic}
\end{algorithm}
Algorithm 2 shows the adjusted feature selection scheme, which includes backward and forward feature selection methods. 
The forward feature selection adds features one by one from the Adjusted feature set $\mathcal{A} $ to Best feature set $\mathcal{B} $. The correlated sensitivity is calculated according to Equation (7), and then Laplace noise is added according to Equation (8).
Training with these added features can obtain the feature set  $\mathcal{A}_1 $, which provides optimal performance. However, sometimes adding a large number of features only slightly increases performance, particularly with high dimension datasets, while too many features can lead to a less interpretive model. Hence, when a set of added features appears to be more or less equally good, then it makes sense to choose the simplest feature set. We set a threshold $ T $ to evaluate the difference of training results. If the difference of training results is smaller than  $ T $, we select the simplest feature set that has the smallest number of predictors.

In backward feature selection, features in set are deleted one by one according to their feature importance. By comparing the training results with different deleted features, feature set $\mathcal{A}_2 $ is generated, which has the best performance. Similar to forward feature selection, when a set of deleted features appears to be more or less equally good, it makes sense to choose the simplest feature set. We also use the threshold $ T $ to select the simplest feature set.
Ultimately, the Adjusted feature set $ \mathcal{A} $  is determined by comparing the training result $s(\mathcal{A}_1)  $ and $s(\mathcal{A}_2 ) $.

\subsection{Discussion}
 Best feature subset $\mathcal{B} $ and Adjusted feature set $\mathcal{A} $,  represent the balance between  
utility and correlated sensitivity. Adding the adjusted features is likely to degrade data utility somewhat, but these extra features serve to reduce the correlated sensitivity of the dataset, which offsets the reduction in utility. The overall result is a feature selection scheme that strikes a balance that leads to less data correlation while maintaining good data utility for data analysis and data publishing.

Our proposed scheme has three advantages.
First, feature importance is a computationally-efficient method for generating the best feature set compared to some of the other existing methods.
Feature importance is the variable that provides the guide to select which features are best to add or delete. Second, with differential privacy, we can choose features privately.
Third, with the consideration of data correlation, we can select features that has less data correlation in the whole dataset and thus reduce the correlated sensitivity  and improve the data utility of the dataset.
\section{Theoretical analysis}
\subsection{Privacy analysis}
\begin{Theo}
	The proposed CR-FS scheme satisfies $ \epsilon $-differential privacy.
\end{Theo}
To prove that the proposed CR-FS scheme is satisfied with differential privacy, we first analyze which steps consume privacy budget in CR-FS scheme. According to Algorithm 1 and Algorithm 2, we access the dataset in two places: 1) the process of feature selection and, 2) the process of data training. To protect the data privacy, we add differential privacy noise in these two places. 

 We split the total privacy budget $ \epsilon $ into two parts  $ \epsilon_{1}$ and $ \epsilon_{2} $ and allocate $ \epsilon_{1}$ and $ \epsilon_{2} $ in the process of feature selection and the process of data training, respectively. First, we analyze the privacy budget $ \epsilon_{1} $ in the process of feature selection.

\begin{lem}
	The process of feature selection satisfies $ \epsilon_{1} $-differential privacy.
\end{lem}

We know that $ D $ and $ D^{'} $ are any two datasets that differ in one feature, and $ f_{1}(\cdot) $ is the query for feature selection.  $ p_{x}(z) $ and $ p_{y}(z) $ denote the probability density function as, 
\begin{equation}
\mathcal{M}_{1}(x,f_{1}(\cdot),\epsilon_{1}) = f_{1}(x) + Lap(\frac{\Delta fim_{q}}{\epsilon_{1}}) \tag{14}
\end{equation}
Let $ x,y $ be two neighboring datasets. We compare two random points $ z \in \mathbb{R} $ and the ratio of two probability density can be presented as

\begin{align} \frac{p_{x}(z)}{p_{y}(z)} \tag{15} &= \prod_{i=1}^{N} \left(\frac{\exp \left(-\frac{\varepsilon_{1}\left|f_{1}(x) i-z_{i}\right|}{\Delta fim_{q}}\right)}{\exp \left(-\frac{\varepsilon_{1}\left|(f_{1}(y)i-z_{i} |\right.}{\Delta fim_{q}}\right)}\right)\\ \notag
&= \prod_{i=1}^{N} \exp \left(\frac{\varepsilon_{1}\left(\left|f_{1}(y)_{i}-z_{i}\right|-\left|f_{1}(x)_{i}-z_{i}\right|\right)}{\Delta fim_{q}}\right)\\ \notag
& \leq \prod_{i=1}^{N} \exp \left(\frac{\varepsilon_{1} | f_{1}(x)_{i}-f_{1}(y)_{i} \|}{\Delta fim_{q}}\right)\\ \notag
&=\exp \left(\frac{\varepsilon_{1} \cdot\|f_{1}(x)-f_{1}(y)\|_{1}}{\Delta fim_{q}}\right) \\ \notag
& \leq \exp (\varepsilon_{1}) \notag
\end{align}
where the first inequality is from triangle inequality and the second inequality is from Equation (11).
The sensitivity of feature selection is according to the maximal difference of feature importance. Therefore, the process of feature selection satisfies $ \epsilon_{1}  $-differential privacy. Second, we analyze the privacy budget $ \epsilon_{2} $ in the process of data training. 
\begin{lem}
	The process of data training satisfies $ \epsilon_{2} $-differential privacy.
\end{lem}
We know that $ D $ and $ D^{'} $ are any two datasets that differ in one record. $ f_{2}(\cdot)$ is the query for training results.The differential privacy noise is added to the weights in training algorithms, such as Linear Regression (LR) and Support Vector Machine (SVM). $ f_{2}(\cdot) $ is the query for the training results. We use $ v_{x}(z) $ and $ v_{y}(z) $ to denote the probability density function as,
\begin{equation}
\mathcal{M}_{2}(x,f_{2}(\cdot),\epsilon_{2}) = f_{2}(x) + Lap(\frac{\Delta CS_{q}}{\epsilon_{2}}) \tag{16}
\end{equation}

The ratio of two probability density can be presented as

\begin{align} \frac{v_{x}(z)}{v_{y}(z)}  \tag{17} &=\prod_{i=1}^{N}  \left(\frac{\exp \left(-\frac{\varepsilon_{2}\left|f_{2}(x) i-z_{i}\right|}{\Delta CS_{q}}\right)}{\exp \left(-\frac{\varepsilon_{2}\left|(f_{2}(y)i-z_{i} |\right.}{\Delta CS_{q}}\right)}\right)\\ \notag
&=\exp  \left(\frac{\varepsilon_{2} \cdot\|f_{2}(x)-f_{2}(y)\|_{1}}{\Delta CS_{q}}\right) \\ \notag
& \leq \exp (\varepsilon_{2}) \notag
\end{align}
 The 	$\Delta CS_{q} =\max \limits_{i \in q} \sum_{j=0}^{l} |\theta_{ij}| \{	 \lVert (Q(D^{j})-Q(D^{-j}) \rVert_{1}    \} $, hence the data training satisfies $ \epsilon_{2} $-differential privacy.

 In the CR-FS scheme, we add privacy budget $ \epsilon_{1} $ and privacy budget $ \epsilon_{2} $ sequentially. Combined with $ \textbf{Lemma1} $, $ \textbf{Lemma2} $ and $ \textbf{Theorem 1} $, we can prove that the proposed  CR-FS  scheme satisfies $ \{\epsilon_{1} + \epsilon_{2}\} $-differential privacy.

 
\section{Experiments}
Our evaluation experiments involve four real-world datasets in terms of both data analysis and data publishing tasks \cite{bibitem Zhu2017}. Utility for data analysis is tested with two machine learning algorithms: LR and linear SVM. Utility for data publishing is tested on count and mean queries. 

\subsection{Experimental setup}
\subsubsection{Dataset}
The experiments involve four datasets, which have different extent of data correlation and different number of features. 
\begin{itemize}
	\item Adult Dataset \cite{bibitem dataset1}: Adult Dataset is from the UCI Machine Learning repository.
	After data preprocessing, we extract 3000 records with 12 features.
	\item Breast cancer Dataset \cite{bibitem dataset2}: This dataset can be found on UCI Machine Learning Repository.
	 After data preprocessing resulted in 569 records with 20 features.
	\item Titanic Dataset \cite{bibitem dataset3}: This dataset comes from a Kaggle competition where the goal was to analyze which sorts of people were likely to survive the sinking of the Titanic. After data preprocessing, we extract 891 records with 9 features.
	\item Porto Seguro Dataset \cite{bibitem dataset4}: Porto Seguro is a well-known auto and homeowner insurance company. 
	After preprocessing, we extract 1770 records with 37 features. 
\end{itemize}

\subsubsection{Comparison}
For better comparisons, four schemes are considered in the experiments.
\begin{itemize}
	\item A non-private scheme, where the dataset has no privacy protection.
	\item The group scheme, where noise is added by multiplying the number of correlated records, as proposed by Chen et al. in \cite{bibitem Chen2014}.
	\item The Zhu scheme, where noise is added according to the correlated sensitivity \cite{bibitem Zhu2015}.
	\item  The proposed scheme, where noise is added according to the CR-FS scheme defined in this paper.
\end{itemize}

\subsubsection{Parameters}
For correlation knowledge between records, no dataset suggests pre-defined knowledge of any correlated data.  We use Pearson correlation coefficient to construct the correlated degree matrix, where a correlation exists for record $i$ and record $j$ if $ \theta_{ij} \geq \theta_0 $. $\theta_0 $ is set to 0.9 for Adult Dataset, Breast cancer Dataset and Breast cancer Dataset and $\theta_0 $ in Porto Seguro Dataset is set to 0.7. For correlation knowledge between features, the Pearson correlation coefficient threshold $ T_{fi} $ is set to 0.9. The missing value threshold $ T_{mv} $ is set to 0.2. The threshold of feature importance $ T_{fi} $ is set to 0.9.

\begin{figure}[ht]
	\centering
	\begin{minipage}[b]{0.49\linewidth}
		\includegraphics[scale=0.24]{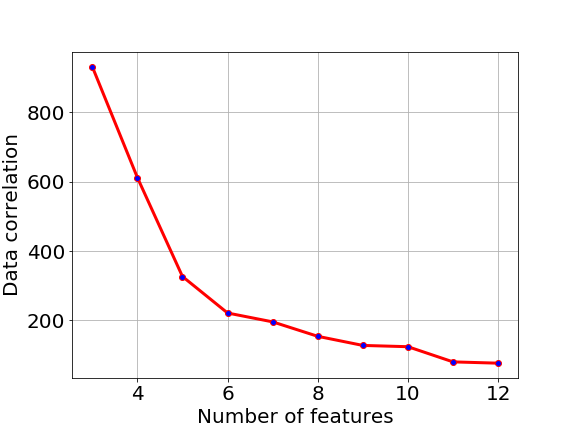}
		\centerline{(a) Adult}
	\end{minipage}
	\begin{minipage}[b]{0.49\linewidth}
		\includegraphics[scale=0.24]{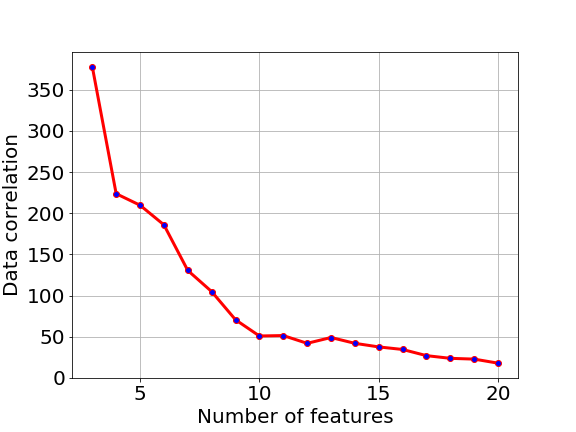}
		\centerline{(b)  Breast Cancer}
	\end{minipage}
	\begin{minipage}[b]{0.49\linewidth}
		\includegraphics[scale=0.24]{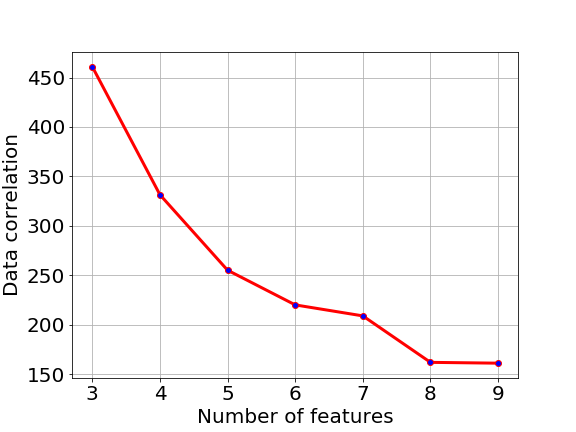}
		\centerline{(c) Titanic}
	\end{minipage}
	\begin{minipage}[b]{0.49\linewidth}
		\includegraphics[scale=0.24]{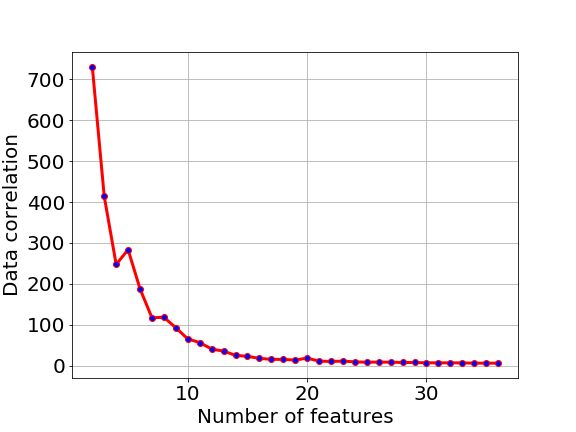}
		\centerline{(d) Porto Seguro}
	\end{minipage}
	\caption{Data correlation for different number of features}\label{fig:1d}
\end{figure}

\begin{table}[htb]
	\centering
	\caption{Number of features in different stages}
	\footnotesize
	\begin{tabular}{lllll}
		\hline
		& \tabincell{c}{Original\\ dataset}  & \tabincell{c}{After data \\preparation} & \tabincell{c}{Best feature \\set $ \mathcal{B} $} & \tabincell{c}{Adjusted feature \\set $ \mathcal{A} $} \\
		\hline
		Adult          & 15                & 12                    &8                              & 12 \\
		Breast cancer   & 32                & 20                    &  10                            &17   \\
		Titanic        & 12                &9                      &7                             &9\\
		Porto seguro    & 59                &37                      &14                            &28\\
		\hline	\end{tabular}
\end{table}

\subsection{Experiments for data analysis}
One aim of our proposed scheme is to improve utility for data analysis, which we evaluate according to the accuracy of the predicted results. For this set of experiments, we choose two machine learning algorithms - LR and linear SVM - and test the output perturbation to assess data utility.

\begin{figure}[ht]
	
	\begin{minipage}[b]{0.49\linewidth}
		\centering	
		\includegraphics[scale=0.24]{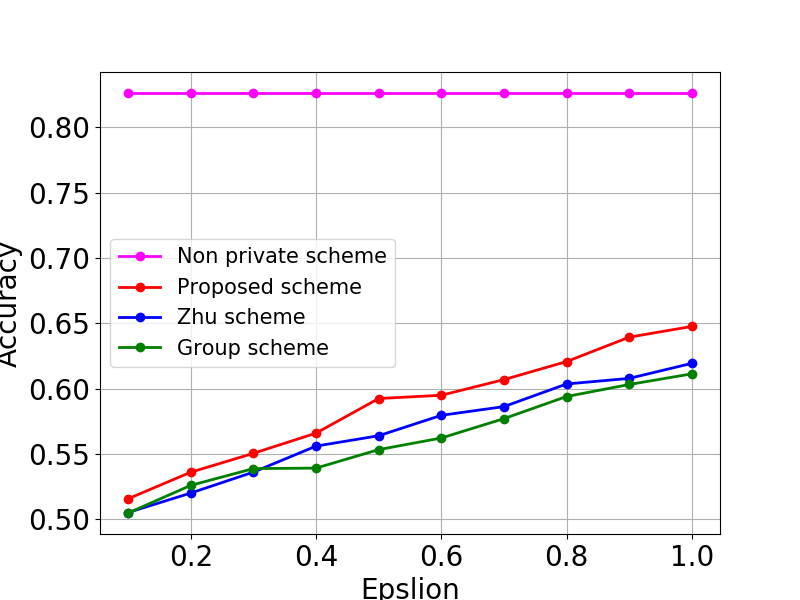}
		\centerline{(a) Adult}
	\end{minipage}
	\begin{minipage}[b]{0.49\linewidth}
		\includegraphics[scale=0.24]{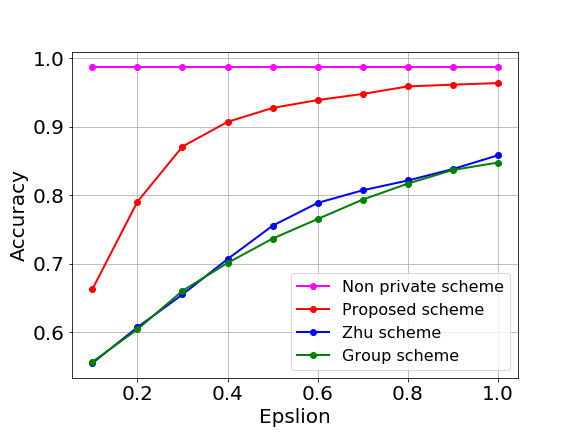}
		\centerline{(b) Breast cancer}
	\end{minipage}
	\begin{minipage}[b]{0.49\linewidth}
		\centering	
		\includegraphics[scale=0.24]{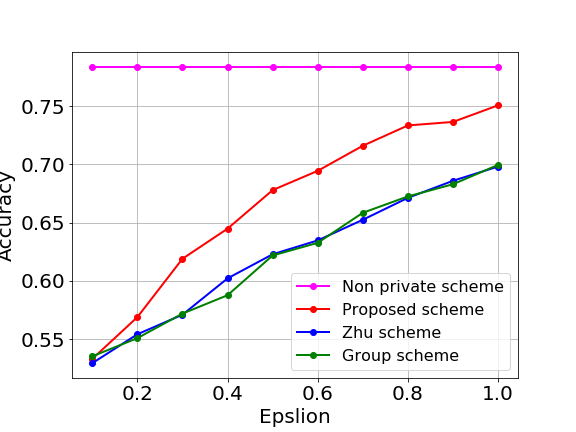}
		\centerline{(c) Titanic} 
	\end{minipage}
	\begin{minipage}[b]{0.49\linewidth}
		\includegraphics[scale=0.24]{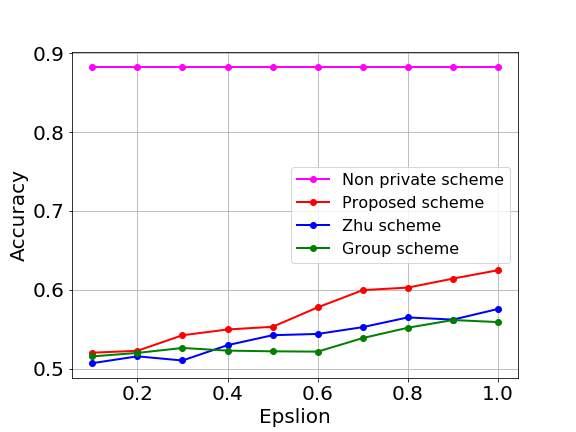}
		\centerline{(d) Porto Seguro}
	\end{minipage}
	\caption{Privacy-Accuracy trade-off in SVM for different datasets}\label{fig:1d}
\end{figure}

Fig. 2 shows that, according to the Pearson correlation coefficients, data correlation varies with the number of features. Data correlation generally decreases with a growing number of features but eventually stabilizes. For example, Figs. 2b and 2c show that data correlation become stable at 17 features with the Breast Cancer dataset and at 8 features with the Titanic dataset. This observation indicates that data correlation across the entire dataset can be reduced while preserving a suitable number of features for data analysis because more features means less correlation.

Table 3 shows the number of features in each dataset at different stages of the proposed scheme. It is noted that, Best feature set $ \mathcal{B} $ will always contain more features than Adjusted feature set  $ \mathcal{A} $ and, as shown in the table, Adjusted feature set $ \mathcal{A} $ have less data correlation than Best feature set $ \mathcal{B} $, demonstrating that more features reduces correlation in a correlated dataset.

Figs. 3 and 4 show the performance of linear SVM and LR on different datasets with the four schemes. 
In most cases, LR have better accuracy than linear SVM. For example, Fig. 2a shows that when   $ \epsilon=1 $, LR have an accuracy of around 0.675 versus linear SVM’s 0.645. Accuracy with the non-private scheme remains constant as the privacy budget increases and also performed better than the other schemes. This result demonstrates that imposing any form of privacy requirement on a dataset degrades data utility.

For the private schemes, the proposed scheme outperforms both the group and Zhu schemes in all circumstances. Figs. 3 and 4 show the level of improvement,  especially Fig. 3b. $ \epsilon=1 $, the proposed scheme scores an accuracy of around 0.97 compared to around 0.85 for the Zhu scheme. We attribute the improved performance of our scheme to the adjusted features. These additional features reduce data correlation but have little impact on the prediction results. Less data correlation means less noise needs to be added, which leads to better data utility. Other schemes do not reduce data correlation; they only consider how to accurately describe the data correlations, without considering that data correlation actually impedes accuracy. 

\begin{figure}[ht]
	
	\begin{minipage}[b]{0.49\linewidth}
		\centering
		\includegraphics[scale=0.24]{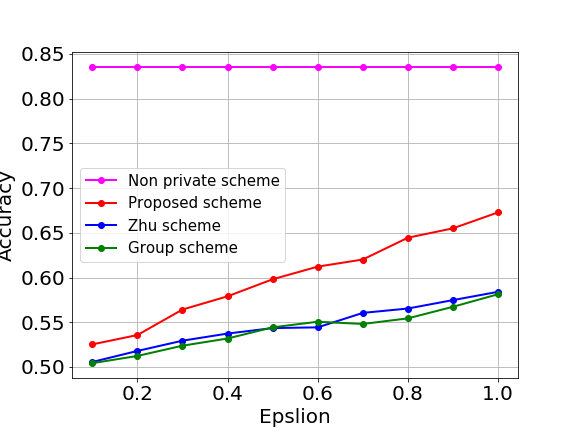}
		\centerline{(a) Adult}
	\end{minipage}
	\begin{minipage}[b]{0.49\linewidth}
		\includegraphics[scale=0.24]{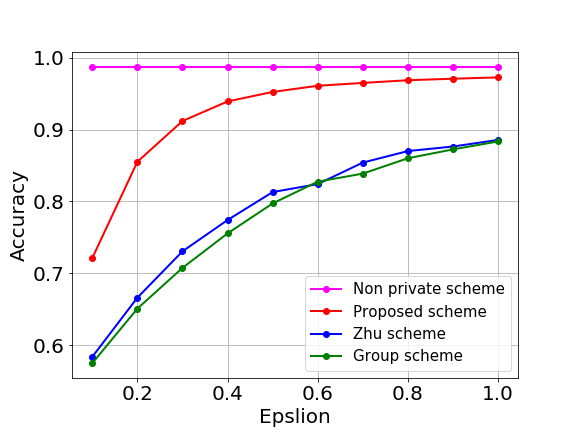}
		\centerline{(b) Breast cancer}
	\end{minipage}
	\begin{minipage}[b]{0.49\linewidth}
		\includegraphics[scale=0.24]{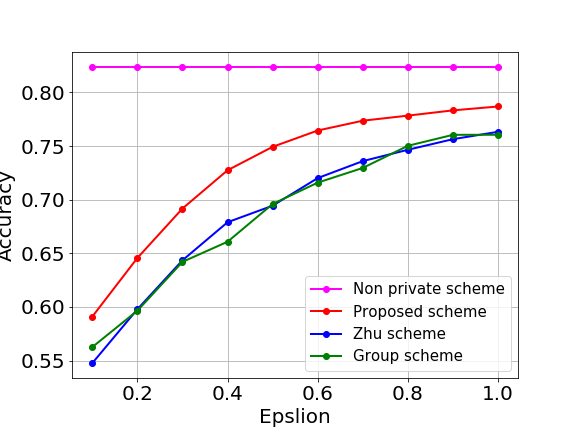}
		\centerline{(c) Titanic}
	\end{minipage}
	\begin{minipage}[b]{0.49\linewidth}
		\includegraphics[scale=0.24]{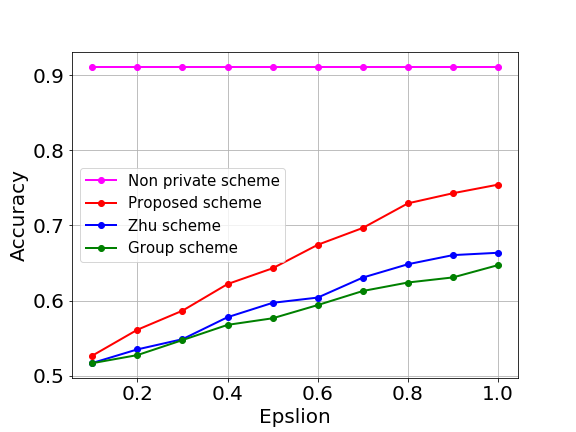}
		\centerline{(d) Porto Seguro}
	\end{minipage}
	\caption{Privacy-Accuracy trade-off in LR for different datasets}\label{fig:1d}
\end{figure}
Additionally, the group and Zhu schemes present closed curves with the first three datasets because the Pearson coefficient is set to a high-value $ \theta_0=0.9 $. This results in a similar correlated sensitivity for both schemes and, consequently, a similar level of noise is added. However, with the Porto Seguro dataset, we set the Pearson coefficient to $ \theta_0=0.7 $. Hence, there is a minor gap in performance. Also worthy of note is that the accuracy of prediction results varied for different datasets. This is due to the amount of data correlation in each dataset; higher correlation means more noise must be added, which reduces accuracy. 

\subsection{Experiments for data publishing}
The second aim of our scheme is to improve utility for data publishing, which we evaluate with both count and mean queries. Mean absolute error (MAE) is used as the metric to assess both results, but different calculation formulas are defined to analyze the base results and the impact of varying the privacy budget. The accuracy  of common queries is measured by MAE, which is given as,
\begin{figure}[ht]
	\centering
	\begin{minipage}[b]{0.49\linewidth}
		\includegraphics[scale=0.24]{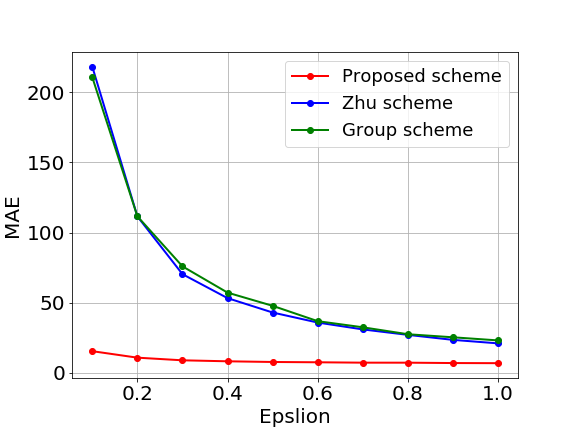}
		\centering
		\centerline{(a) Adult}
	\end{minipage}
	\begin{minipage}[b]{0.49\linewidth}
		\includegraphics[scale=0.22]{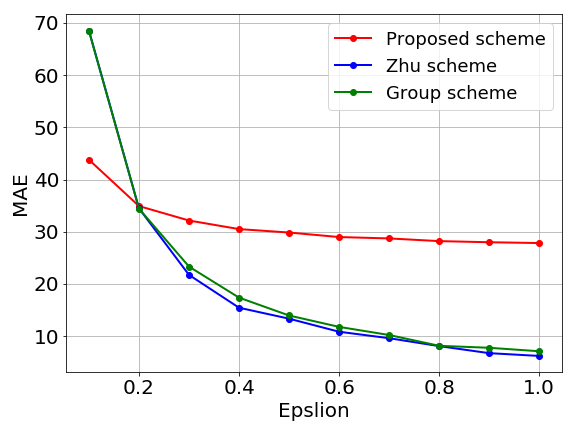}
		\centering
		\centerline{(b) Breast cancer}
	\end{minipage}
	\begin{minipage}[b]{0.49\linewidth}
		\includegraphics[scale=0.24]{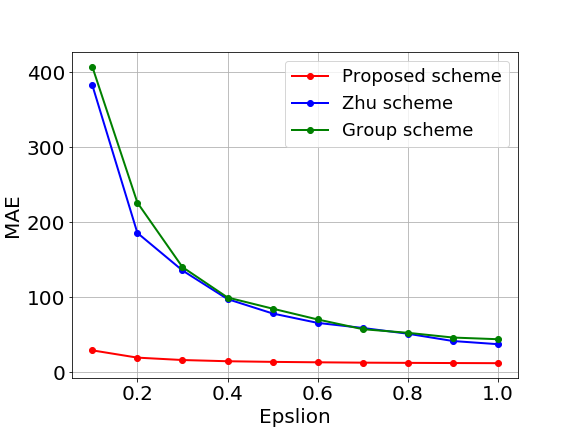}
		\centering
		\centerline{(c) Titanic}
	\end{minipage}
	\begin{minipage}[b]{0.49\linewidth}
		\includegraphics[scale=0.24]{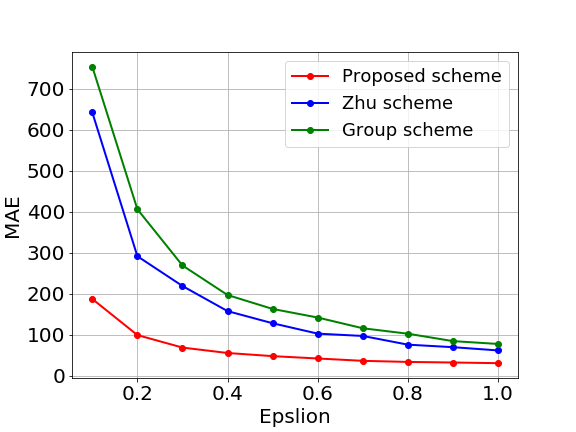}
		\centering
		\centerline{(d) Porto Seguro}
	\end{minipage}
	\caption{MAE performance for count queries}\label{fig:1d}
\end{figure}
\begin{equation}
MAE=\frac{1} {|\mathcal{Q}|} \sum_{\mathcal{Q}_{i} \in \mathcal{Q}}   |\hat{\mathcal{Q}_{i}}(x)- \mathcal{Q}_{i}(x)|
\tag{18}\end{equation}
where $ \mathcal{Q}_{i}(x) $ is the true aggregation result for one query,  and $ \hat{\mathcal{Q}_{i}}(x) $is the perturbed aggregation result calculates through different schemes. A low MAE indicates a low error and, thus, a better data utility.

To analyze how the proposed scheme performs with different privacy budgets, we also define a second MAE containing two components. One component measures the noise added due to correlated sensitivity, the other measures the errors introduced by adding the adjusted features. These features have an impact on a new query object that can emerge as errors when comparing the adjusted dataset to the original. This MAE is defined as

\begin{equation}
MAE=\frac{1} {|\mathcal{Q}|} \sum_{\mathcal{Q}_{i} \in \mathcal{Q}}   |\hat{\mathcal{Q}_{i}}(x)- (\mathcal{Q}_{i}(x)-\mathcal{Q}_{i}^o(x))|
\tag{19}\end{equation}
where $\hat{\mathcal{Q}_{i}}(x)$ and $ \mathcal{Q}_{i}^o(x) $ are the true aggregation result on Best feature set $\mathcal{B}$ and Adjust feature set $\mathcal{A}$, respectively.
\begin{figure}[ht]
	\centering
	\begin{minipage}[b]{0.49\linewidth}
		\centering
		\includegraphics[scale=0.24]{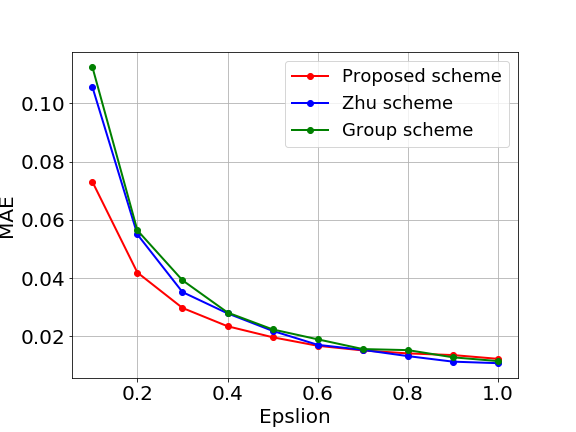}
		\centerline{(a) Adult}
	\end{minipage}
	\begin{minipage}[b]{0.49\linewidth}
		\centering
		\includegraphics[scale=0.24]{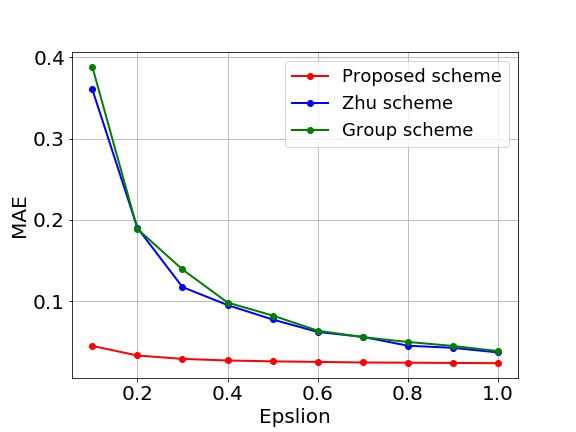}
		\centerline{(b) Breast cancer}
	\end{minipage}
	\begin{minipage}[b]{0.49\linewidth}
		\centering
		\includegraphics[scale=0.24]{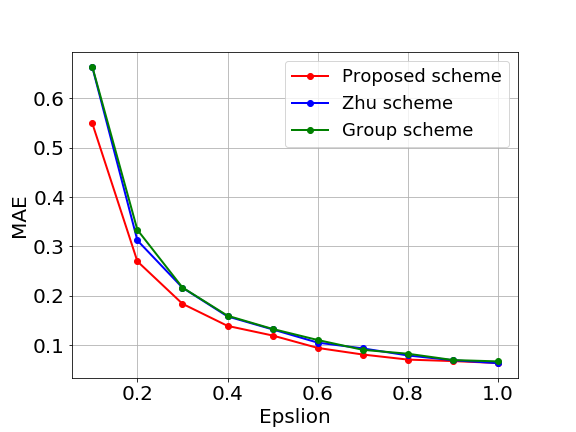}
		\centerline{(c) Titanic}
		
	\end{minipage}
	\begin{minipage}[b]{0.49\linewidth}
		\centering
		\includegraphics[scale=0.24]{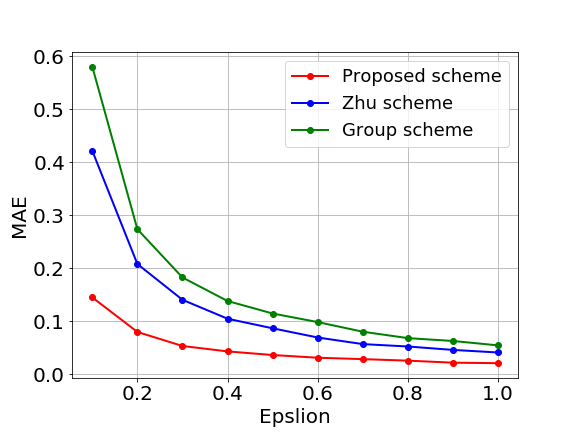}
		\centerline{(d) Porto Seguro}
		
	\end{minipage}
	\caption{MAE performance for mean queries} 
\end{figure}

Fig. 5 shows the impact of varying privacy budgets on the performance of count queries in terms of MAE. With the proposed scheme, the MAE decreases as the privacy budget grows before stabilizing toward the end. This result demonstrates that a lower privacy requirement has better data utility. Moreover, the MAE for the proposed scheme is significantly smaller than the other schemes, which means that the proposed scheme does indeed improve data utility. For example, Figs. 5a and 5b at $ \epsilon =0.2 $ show an MAE of around 18 for the Adult dataset and 17 for the Titanic dataset using our proposed scheme, whereas the group and Zhu schemes return an MAE of around 110 and 200 on these same datasets - an enormous increase over the proposed scheme. Again, we attribute these results to reduced data correlation after adding the adjusted features. 

In terms of the other schemes, the MAE for the Zhu scheme is slightly lower than for the group scheme most of the time for the same reason as explained in the data analysis experiments. Moreover, the MAE for the Zhu scheme decreases faster as the privacy budget increased from 0.1 to 0.4 than when the budget increases from 0.4 to 1. This again shows that a higher privacy requirement creates a higher data utility cost.

The results of varying the privacy budgets with mean queries are similar, as shown in Fig. 6. However, the MAE are much smaller than for the count queries. This is because, after data normalization, the scale of data falls within $ [-1,1] $; therefore, each record has a similar mean value. As a result, the outcomes of mean queries are much smaller than for count queries. In addition, the MAE for our proposed scheme is not always better than the group or Zhu schemes - for example, when  $ \epsilon <0.2 $. This shows that adding the adjusted features can introduce additional errors. Hence, the quality of the query results in the proposed scheme depends on the type of queries and the dataset itself but, overall, our proposed scheme returns a lower MAE than the other schemes.

\subsection{Discussion}
The key to the CR-FS scheme is to reduce data correlation in the whole dataset, while maintaining a good utility for data analysis and data queries. We add differential private noise on two places: feature selection and data training and still can achieve desirable performance. This is because the fact that sensitivity of feature selection is smaller than 1, the sequence of feature importance will not change much. That is to say, there is a high probability that more important features are still more important and less important features are still less important. In this way, a higher probability that important features are kept for training and less important features are used to reduce data correlation.

For data analysis, we select features in the step 5 according to the accuracy of predicted results, thus the selected features can have less correlation across the whole dataset and achieve a desirable accuracy results. For data queries, the correlation in the whole dataset is also reduced with the proposed CR-FS scheme. However, as we noted in the Figure 5 and 6, the MAE is not always better than other schemes. This is because the sensitivity is related the type of queries and dataset itself. Deleted or added features in the dataset can reduce the data correlation, which may bring in new error with regard to different queries.

\section{Conclusion }
In this paper, we identified the privacy issue of the data correlation in machine learning, which may result in more privacy loss than expected in industrial applications.
We propose a novel feature selection scheme CR-FS to reduce data correlation with little compromise to data utility. The proposed CR-FS scheme includes steps that consider the accuracy of predicted results, the privacy preserving and the data correlation in the dataset. Our proposed algorithm strikes a better trade-off between data utility and privacy leaks for correlated datasets. The method’s performance is evaluated via extensive experiments, and the results prove that our proposed CR-FS scheme provides better data utility for both data analysis and data queries compared to traditional schemes.

\appendices


\ifCLASSOPTIONcaptionsoff
  \newpage
\fi



%
\section*{Acknowledgment}
This work was supported by National Natural Science Foundation of China 61972366, in part by the Australian Research Council under Linkage Grant LP170100123, and by the ministry of education, humanities, and social science project of China under 19A 10520035.

%

\begin{IEEEbiography}[{\includegraphics[width=1in,height=1.25in,clip,keepaspectratio]{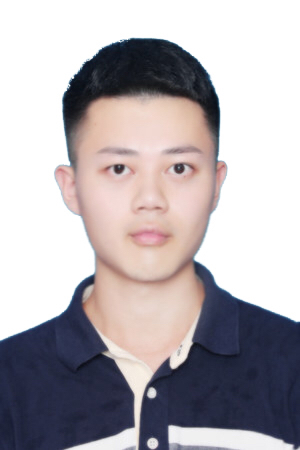}}]{Tao Zhang}
	works towards his Ph.D degree with the school of Computer Science in the University of Technology Sydney, Australia.
	His research interests include privacy preserving, algorithmic fairness, and machine learning.
\end{IEEEbiography}

\begin{IEEEbiography}[{\includegraphics[width=1in,height=1.25in,clip,keepaspectratio]{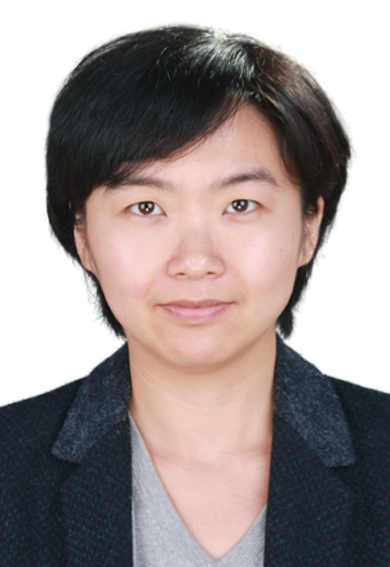}}]{Tianqing Zhu} 
	received her BEng and MEng degrees from Wuhan University, China, in 2000 and 2004, respectively, and a PhD degree from Deakin University in Computer Science, Australia, in 2014. Dr Tianqing Zhu is currently a senior lecturer in the school of Computer Science in the University of Technology Sydney, Australia.  Before that, she was a lecture in the School of Information Technology, Deakin University, Australia, from 2014 to 2018. Her research interests include privacy preserving, data mining and network security. 
	
\end{IEEEbiography}

\begin{IEEEbiography}[{\includegraphics[width=1in,height=1.25in,clip,keepaspectratio]{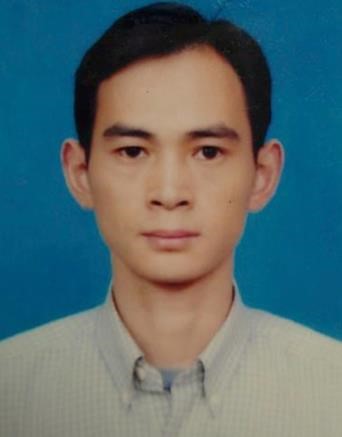}}]{Ping Xiong}
	received his BEng degree from LanZhou Jiaotong University, China in 1997. He received his MEng and PhD degrees from Wuhan University, China, in 2002 and 2005, respectively. 
	He is currently the professor of School of Information and Security Engineering, Zhongnan University of Economics and Law, China. His research interests are network security, data mining and privacy preservation.
	
\end{IEEEbiography}

\begin{IEEEbiography} [{\includegraphics[width=1in,height=1.25in,clip,keepaspectratio]{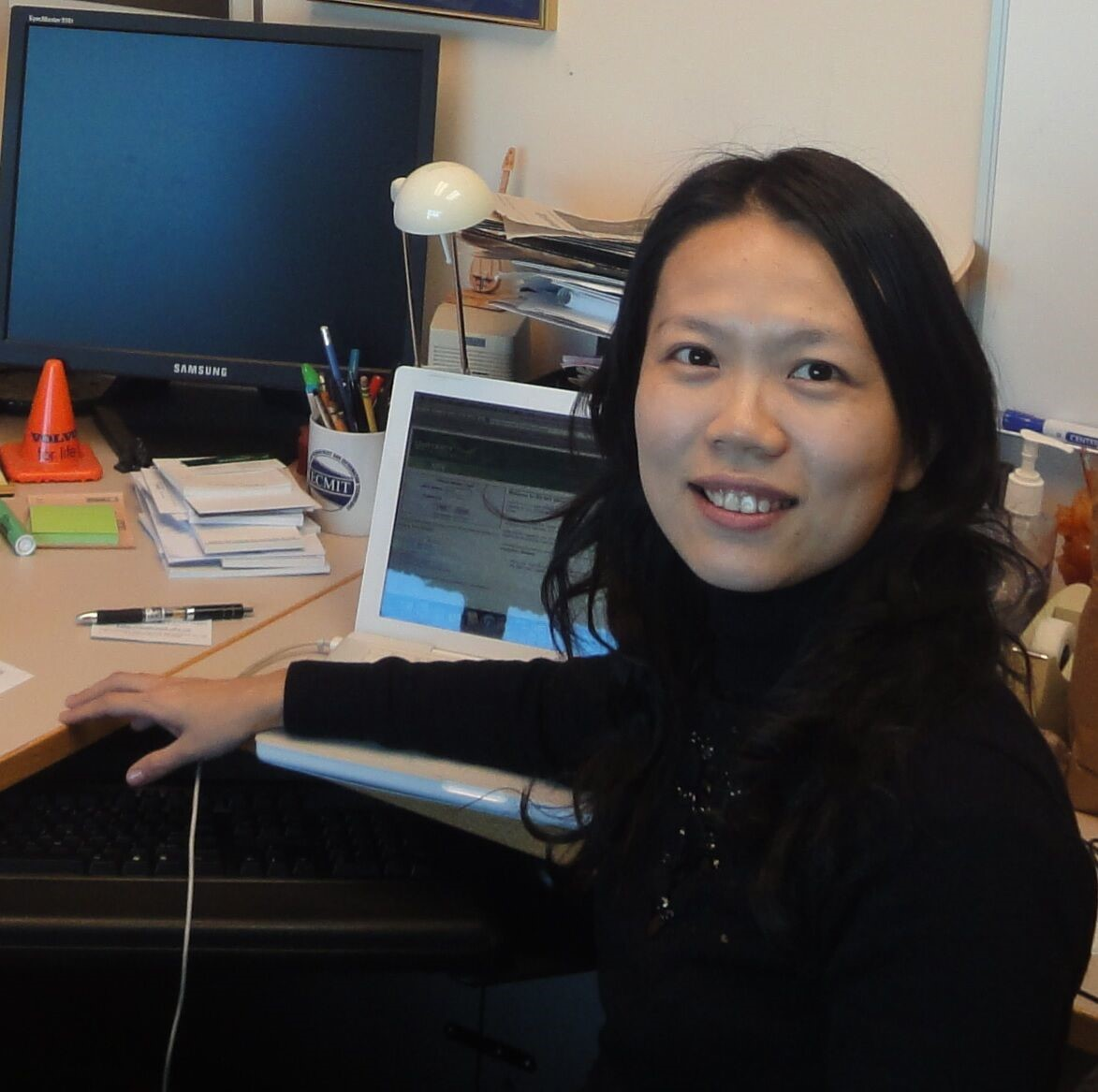}}]{Huan Huo}
	received the B.Eng and Ph.D. degrees from Northeastern University, China in 2002 and 2007, both in Computer Science and Technology. 
	From 2012 to 2014, Angela HUO taught at the Department of Computer Information System, the University of the Fraser Valley in Canada, and did collaborative research in the University of Waterloo as a visiting scholar for one year. Since 2018, she has been a senior lecture in the school of Computer Science at the University of Technology Sydney, Australia. Her research interests include data stream management technology, advanced data analysis, and data-driven cybersecurity.
	
\end{IEEEbiography}

\begin{IEEEbiography}[{\includegraphics[width=1in,height=1.25in,clip,keepaspectratio]{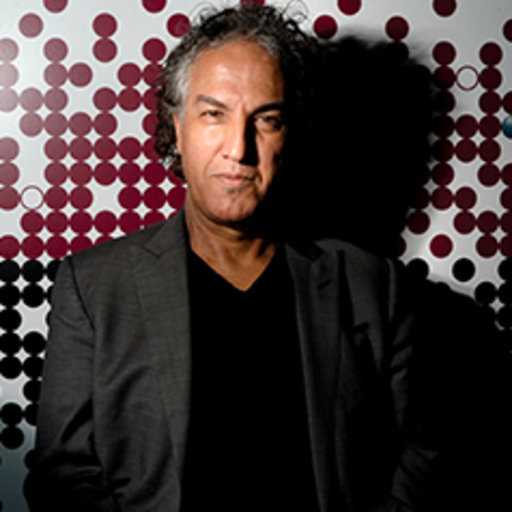}}]{Zahir Tari}
		is a full professor in distributed systems at RMIT University, Australia. He received his Ph.D. degree in computer science from the University of Grenoble, France, in 1989. His expertise is in the areas of system performance (e.g., cloud, IoT) as well as system security (e.g., SCADA, cloud). He was an Associate Editor of IEEE Transactions on Computers, IEEE Transactions on Parallel and Distributed Systems, and IEEE Magazine on Cloud Computing.
\end{IEEEbiography}

\begin{IEEEbiography}[{\includegraphics[width=1in,height=1.25in,clip,keepaspectratio]{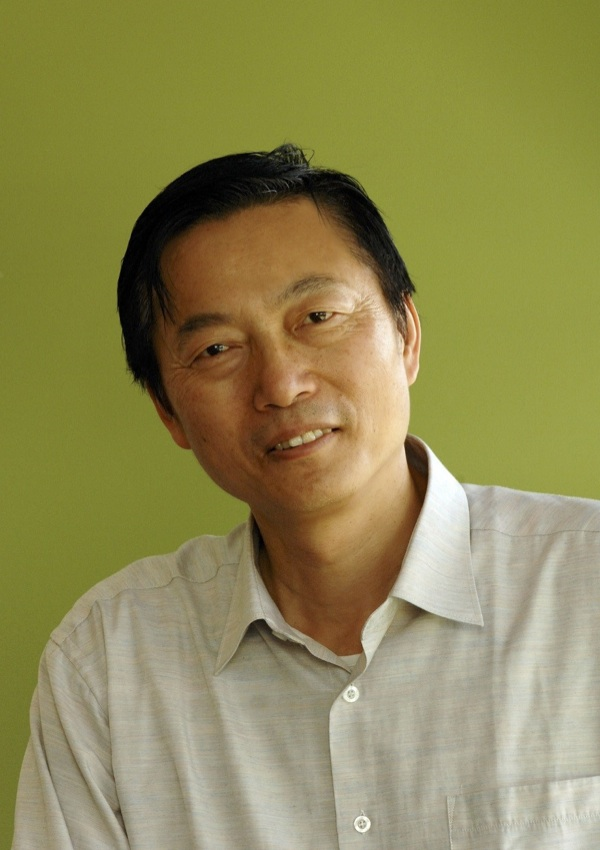}}]{Wanlei Zhou}
	received the B.Eng and M.Eng degrees from Harbin Institute of Technology, Harbin, China in 1982 and 1984, respectively, and the PhD degree from The Australian National University, Canberra, Australia, in 1991, all in Computer Science and Engineering. He also received a DSc degree (a higher Doctorate degree) from Deakin University in 2002. He is currently the Head of school of Computer Science in University of Technology Sydney (UTS). Before joining UTS, Professor Zhou held the positions of Alfred Deakin Professor, Chair of Information Technology, and Associate Dean (International Research Engagement) of Faculty of Science, Engineering and Built Environment, Deakin University. His research interests include security and privacy, bioinformatics, and e-learning. Professor Zhou has published more than 400 papers in refereed international journals and refereed international conferences proceedings, including many articles in IEEE transactions and journals.
\end{IEEEbiography}




\end{document}